\documentclass{article} 
\usepackage{iclr2021_conference,times}

\usepackage{amsmath,amsfonts,bm}

\def\eqref#1{equation~\ref{#1}}

\def\1{\bm{1}}

\DeclareMathAlphabet{\mathsfit}{\encodingdefault}{\sfdefault}{m}{sl}
\SetMathAlphabet{\mathsfit}{bold}{\encodingdefault}{\sfdefault}{bx}{n}

\DeclareMathOperator*{\argmax}{arg\,max}

\usepackage{hyperref}
\usepackage{url}            
\usepackage{booktabs}       
\usepackage{amsfonts}       
\usepackage{nicefrac}       
\usepackage{microtype}      
\usepackage{amsmath}
\usepackage{amsthm}
\usepackage[ruled,vlined]{algorithm2e}
\usepackage{tikz}
\usetikzlibrary{bayesnet}
\usepackage{soul}

\SetKwInput{KwData}{The data}
\SetKwInput{KwResult}{The result}

\newtheorem{theorem}{Property}
\newtheorem*{defn}{Definition}
\newtheorem{prop}{Proposition}
\newtheorem{remark}{Remark}

\newcommand{\decod}{\mathrm{Dec}}

\title{Quasi-symplectic Langevin Variational Autoencoder}

\author{Zihao WANG \thanks{Corresponding author.} \\
Inria Sophia Antipolis, University Côte d’Azur\\
2004 Route des Lucioles, 06902 Valbonne\\
\texttt{zihao.wang@inria.fr} \\
\And
Hervé Delingette \\
Inria Sophia Antipolis, University Côte d’Azur \\
2004 Route des Lucioles, 06902 Valbonne\\
\texttt{herve.delingette@inria.fr }
}

\begin{document}

\maketitle

\begin{abstract}
Variational autoencoder (VAE) is a very popular and well-investigated generative model in neural learning research. To leverage VAE in practical tasks dealing with a massive dataset of large dimensions, it is required to deal with the difficulty of building low variance evidence lower bounds (ELBO). Markov Chain Monte Carlo (MCMC) is an effective approach to tighten the ELBO for approximating the posterior distribution and Hamiltonian Variational Autoencoder (HVAE) is an effective MCMC inspired approach for constructing a low-variance ELBO that is amenable to the reparameterization trick. The HVAE adapted the Hamiltonian dynamic flow into variational inference that significantly improves the performance of the posterior estimation. We propose in this work a Langevin dynamic flow-based inference approach by incorporating the gradients information in the inference process through the Langevin dynamic which is a kind of MCMC based method similar to HVAE. Specifically, we employ a quasi-symplectic integrator to cope with the prohibit problem of the Hessian computing in naive Langevin flow. We show the theoretical and practical effectiveness of the proposed framework with other gradient flow-based methods.


\end{abstract}

\section{Introduction}
\label{introduction}
Variational Autoencoders (VAE) are a popular generative neural model applied in a vast number of  practical cases to perform unsupervised analysis and to generate specific dataset. It has the advantages of offering a quantitative assessment of generated model quality and being less cumbersome to train compared to Generative Adversarial Networks (GANs). The key factor influencing the performance of VAE models is the quality of the marginal likelihood approximation in the corresponding evidence lower bound (ELBO). 

A common method to make the amortized inference efficient is to constraint the posterior distribution of the latent variables to follow a given closed-form distribution, often multivariate Gaussian~\citep{wolf2016variational}. However, this severely limits the flexibility of the encoder. In~\citep{salimans15}, the Hamiltonian Variational Inference (HVI) is proposed to remove the requirement of an explicit formulation of the  posterior distribution by forward sampling a Markov chain based on Hamiltonian dynamics. It can be seen as a type of normalizing flows (NFs) \citep{rezende15Flow} where repeated transformations of probability densities are replaced by time integration of space and momentum variables. 
To guarantee the convergence of HVI to the true posterior distribution, \citeauthor{wolf2016variational} proposed to add an acceptance step in HVI algorithm. Further more, \citet{Caterini2018HamiltonianVA} first combined VAE and HVI in Hamiltonian Variational Autoencoders (HVAE) which include a dynamic phase space where momentum component $\rho$ and position component $z$ are integrated. The using of Hamiltonian flow for the latent distribution inference can introduce the target information (gradient flow) into the inference steps for improving the variational inference efficiency. 

In this work, as an exploration of the application of dynamic systems in the field of machine learning, we propose a novel inference framework named quasi-symplectic Langevin variational auto-encoder (Langevin-VAE) that leads to both reversible Markov kernels and phase quasi-volume invariance similarly to Hamiltonian flow~\citep{Caterini2018HamiltonianVA} while reducing the computation and memory requirements. 
The proposed method is a low-variance unbiased lower bound estimator for infinitesimal discretization steps but needs  just one target Jacobian calculation and avoids computing the Hessian of Jacobian. The leapfrog integrator of Hamiltonian flows is replaced in our approach by  the quasi-symplectic Langevin integration. 
We show that the Langevin-VAE is a generalized stochastic inference framework since the proposed Langevin-VAE becomes symplectic  when the viscosity coefficient $\nu = 0$ is set to zero.   The method is verified through quantitative and qualitative comparison with conventional VAE and Hamiltonian-VAE inference framework on a benchmark dataset.
\section{Preliminary}
\subsection{Variational Inference and Normalizing Flow}
One core problem in the Variational Inference (VI) task is to find a suitable replacement distributions $q_\theta(z)$ of the posterior distribution $p(z|x)$ for optimizing the ELBO: $\operatorname*{arg max}_\theta \mathbb{E}_{q}[\log{p(x, z)} - \log{q_\theta(z)}]$.
To tackle this problem, \citeauthor{Ranganath2014BlackBV} proposed black box variational inference by estimating the noisy unbiased gradient of ELBO to perform direct stochastic optimization of ELBO. \citeauthor{KingmaW13} proposed to use some multivariate Gaussian posterior distributions 
of latent variable $z$ generated by a universal function $\omega$, which makes reparameterization trick is possible:~$\operatorname*{arg max}_{\theta,\phi} \mathbb{E}_{q_{\theta}(z|x)}[\log{p_\theta(x, z)} - \log{q_\phi(z|x)}]$.
To better approximate potentially complex  posterior distributions of latent variables, the use of simple parametric distributions like multivariate Gaussian is a limitation.
Yet only a few of distributions are compatible with the reparameterization trick. Normalizing Flows (NFs)~\citeauthor{rezende15Flow} was proposed as a way to deal with more general parametric posterior distributions that  can still be efficiently optimized with amortized inference 
~\citep{Papamakarios}. NFs are a 
 class of methods that use a series of invertible transformation $T_I...\circ ...T_0$  to map a simple  distribution $z_0$ into a  complex one $z_i$: $z_i = T_I...\circ ...T_0(z_0)$,
By applying a cascade of transformations, the corresponding logarithm prior probability $p(z_i)$ of the transformed distribution becomes:
\begin{equation}
\log(p(z_i)) = \log(p(z_0)) - \Sigma_0^i \log\left|det \frac{\partial T_i}{\partial z_{i-1}}\right|
\label{eq:NFsF}
\end{equation}
where the non-zero Jacobian $|det \frac{\partial T_i}{\partial z_{i-1}}|$ of each transformation ensures the global volume invariance of the probability density.
The positivity of each Jacobian terms is guaranteed by the invertibility of each transformation $T$ and consequently by the 
reversibility of normalizing flows. 

The Hamiltonian dynamics in HVAE can also be seen as a type of NFs, for which Eq:~(\ref{eq:NFsF}) also holds. Briefly, HVAE employs an I steps Hamiltonian: $\mathcal{H}_I$ transformation process to build an unbiased estimation of posterior $q(z)$ by extending $\tilde{p}(x,z)$ as $\tilde{p}(x, \mathcal{H}_I(z_{0},\rho_{0}))$  leading to: $\tilde{p}(x):= \frac{\hat{p}(x, \mathcal{H}_I(z_{0},\rho_{0}))}{q(\mathcal{H}_I(z_{0}, \rho_{0}))}$, where: $\hat{p}(x,z_{I},\rho_{I}) = \hat{p}(x, \mathcal{H}_I(z_{0},\rho_{0})) = \hat{p}(x, z_{I})\mathcal{N}(\rho_{I}|0,I)$.
HVAE in particular, enables the use of the  reparameterization trick during inference thus leading to efficient ELBO gradients computation. 
The Hamiltonian dynamics is such that the distribution of phase space $z,\rho$ remains constant along each trajectory according to Liouville's theorem (symplectic)~\citep{Liouville}. When using the leapfrog integrator with step size $l$ for discretizing the Hamiltonian dynamics, the Jacobian remains to 1 (ignoring numerical rounding errors) with 
$\lim_{l\to 0}{|det \frac{\partial T_i}{\partial z_i}|_l^{-1}} = 1$. This property simplifies the Jacobian calculations  at each discretization step~\citep{Caterini2018HamiltonianVA}.  In HVAE, the posterior approximation is constructed by applying $I$ steps of the Hamiltonian flow:
$q^{I}(\mathcal{H}_i(\theta_{0},\rho_{0}))=q^{0}(\mathcal{H}_i(\theta_{0},\rho_{0}))\prod_{i=1}^{I}|det \nabla{\Phi^{i}(\mathcal{H}_i(\theta_{0},\rho_{0}))|^{-1}}$, where $\Phi^{i}$ represents the leapfrog discretization transform of Hamiltonian dynamics. When  combined with the reparameterization trick, it allows to compute an unbiased estimator of the lower bound gradients $\nabla_{\theta} \mathbb{L}$.
\subsection{Langevin Monte-Carlo and Normalizing Flow}


A Langevin dynamics describes a stochastic evolution of particles within the 
particle interaction potential $U(x)$ that can be treated as a log probability density, it has recently attracted a lot of attention in the machine learning community~\citep{stuart2004,Girolami,Welling2011BayesianLV,mou2020highorder} for the stochastic sampling of posterior distributions $p_{\Phi}(z|x)$ in Bayesian inference. Langevin Monte-Carlo methods~\citep{Girolami} rely on the construction of Markov chains with stochastic paths parameterized by $\Phi$  based on the discretization of the following  \textit{Langevin–Smoluchowski}
 SDE~\citep{Girolami} related to the overdamped Langevin dynamics~:
\begin{equation}
\delta \Phi(t) = \frac{1}{2} \nabla_{\Phi} \log(p(x,\Phi))\delta t + \delta \sigma(t)
\label{langevinDynamic}
\end{equation}
where $\sigma(t)$ is govern by the fluctuation–dissipation theorem \footnote{https://en.wikipedia.org/wiki/Fluctuation-dissipation\_theorem} and $t$ represents the time step.
The stochastic flow in Eq~(\ref{langevinDynamic}) can be further exploited to construct  Langevin dynamics based normalizing flow and its derived methods for posterior inference~\citep{wolf2016variational, nfIntroduce}. The concept of Langevin normalizing flow was first briefly sketched by \cite{rezende15Flow} in their seminal work. 
To the best of our knowledge,  little work has explored practical implementations of Langevin normalizing flows. In \citep{gu}, the authors  proposed a Langevin normalizing flow where invertible mappings are based on overdamped Langevin dynamics discretized with the Euler–Maruyama scheme. The explicit computation of  the Jacobians  of those mappings involves  the  Hessian matrix of $\log(p_{\Phi}(x))$ as follows : 
\begin{equation}
\centering
\begin{split}
\log\left|det \frac{\partial T_i}{\partial z_{k-1}}\right|^{-1} \sim \nabla_z \nabla_z \log(p(x,z)) + \mathbb{O}(z)
\end{split}
\label{eq:hessian}
\end{equation}
Yet, the Hessian matrix appearing in Eq~(\ref{eq:hessian}) is  expensive to compute  both in space and time and adds a significant overhead to the already massive computation of  gradients.
This makes the method of \citep{gu} fairly unsuitable for the inference of complex models. 
In a more generic view, in the Langevin flow, the forward transform is modelled by the Fokker-Plank equation and the backward transform is given by Kolmogorov’s backward equation which is discussed in the work of \citeauthor{nfIntroduce} and is not detailed here.

\subsection{Quasi-symplectic Langevin and Corresponding Flow}
\subsubsection{Trivial Jacobian by Quasi-symplectic Langevin Transform}
To avoid the computation of Hessian matrices in Langevin normalizing flows, we propose to revert to generalized Langevin dynamic process as proposed in \citep{GeneralizedLangevin}. It involves second order dynamics with inertial and damping terms:
\begin{equation}
\begin{split}
    &\delta\Phi(t) = K \delta t \\
    &\delta K(t) = - \frac{\partial ln( p(x,\Phi))}{\partial \Phi}\delta t - \nu K(t) + \delta \sigma(t)
\end{split}
\label{qslangevin}
\end{equation} 
where $\Phi(t)$ and $K(t)$ are the stochastic position and velocity fields, and $\nu$ controls the amount of damping. 
We can see that the Langevin–Smoluchowski type SDE of Eq.:(\ref{langevinDynamic}) is nothing but the special case of high friction motion \citep{GeneralizedLangevin} when Eq.:~(\ref{qslangevin}) has an over-damped frictional force (proof is in \ref{sec;overdamp}).





To get simple Jacobian expressions when constructing Langevin flow, we need to have a symplectic Langevin transformation kernel. To this end, we 
introduce a quasi-symplectic Langevin method for building the flow~\citep{Milstein}. The quasi-symplectic Langevin differs from the Euler–Maruyama integrator method which diverges  for the discretization of generalized  Langevin SDE. Instead, 
the quasi-symplectic Langevin method makes the computation of the  Jacobian tractable during the diffusion process and keeps approximate symplectic properties for the damping  and external potential terms.

More precisely, the quasi-symplectic Langevin integrator is based on the two state variables $(K_i,\Phi_{i})$ that are evolving according to the  mapping $\Psi_\sigma(K_i,\Phi_{i})=(K_{i+1},\Phi_{i+1})$ where $\sigma$ is the kernel stochastic factor. It is known as the \textsl{second order strong quasi-symplectic} method (\ref{eq:quasitransform}) and is composed of the following steps for a time step $t$:
\begin{equation}
\begin{split}
& K_{II}(t, \phi) = \phi e^{-\nu t}\\
    &K_{1,i} = K_{II}(\frac{t}{2}, K_i);~~\Phi_{1,i} = \Phi_i - \frac{t}{2} K_{1,i}\\
    &K_{2,i} = K_{1,i} + t \frac{\partial \log( p(x,\Phi_{1,i}))}{ \partial \Phi_{1,i}} +  \sqrt{t} \sigma \xi_i;~~\xi_i \sim N(0, I)\\
    &K_{i+1} = K_{II}(\frac{t}{2},K_{2,i});~~\Phi_{i+1} = \Phi_{1,i} + \frac{t}{2} K_{2,i}
\end{split}
\label{eq:quasitransform}
\end{equation}
where initial conditions are $K_0 = \kappa_0; \Phi_0 = \phi_0$.  

The above quasi-symplectic integrator satisfies the following two properties:
\begin{theorem}
\label{theorem1}
Quasi-symplectic method degenerates to a symplectic method when $\nu = 0$.
\end{theorem}
\begin{theorem}
\label{theorem2}
Quasi-symplectic Langevin transform $\Psi_0(K_i,\Phi_{i})$ (\ref{eq:quasitransform}) has constant Jacobian  :
\begin{equation}
    |\Psi_0(K_i,\Phi_{i})| = \frac{\partial \Phi_{i+1}}{\partial \Phi_{i}} \frac{\partial K_{i+1}}{\partial K_{i}} - \frac{\partial \Phi_{i+1}}{\partial K_{i}} \frac{\partial K_{i+1}}{\partial \Phi_{i}} = exp(-\nu t)
\end{equation}
\end{theorem}
The first property shows that the VAE constructed based on the Quasi-Symplectic Langevin (QSL) dynamics is conceptually equivalent to a HVAE in the absence of damping $\nu=0$. The second property  implies that the QSL integrator leads to  transformation kernels  that are reversible and with trivial Jacobians. 
The proofs of those two properties can be found in appendix \ref{sec;symplectic} and more discussion about the quasi-symplectic integrator can be found in \cite{Milstein2}.

The advantage of the QSL flow compared to the regular overdamped Langevin flow is that it avoids computing the Hessian of the log probability to compute the Jacobian  which is a major advantage given the complexity of the Hessian computation. 

We give below the formal definition of the quasi-symplectic Langevin normalizing flow.
\label{NFsF}
\begin{defn}
An $I$ steps discrete quasi-symplectic Langevin normalizing flow $\mathcal{L}^{I}$ is defined by a series of diffeomorphism, bijective and invertible mapping $\Psi_0: \sigma_{\mathcal{A}} \xrightarrow{} \sigma_{\mathcal{B}}$ between two measurable spaces $(\mathcal{A}, \sigma_{\mathcal{A}}, \mu_\alpha)$ and $(\mathcal{B}, \sigma_{\mathcal{B}}, \mu_\beta)$:
\begin{equation}
\begin{split}
    \mathcal{L}^I \mu_\alpha(\mathcal{S_{\mathcal{A}}}): &\Psi_i \circ \mu_\alpha (\mathcal{S_{\mathcal{A}}}) = \mu_\alpha (\Psi_{i-1}^{-1}(\mathcal{S_{\mathcal{B}}})) , \\
    &\forall \mathcal{S}_{\mathcal{A}} \in \sigma_{\mathcal{A}},  \mathcal{S}_{\mathcal{B}} \in \sigma_{\mathcal{B}}, i = \{1, ..., I\}.
\end{split}
\end{equation}
\label{defflow}
where $\sigma_{\mathcal{(\cdot)}}$ and $\mu_{(\cdot)}$ are the $\sigma$-algebra and probability measure for set $\mathcal{(\cdot)}$ respectively, $\Psi_i$ is the $i_{th}$ quasi-symplectic Langevin transform given by Eqs:(\ref{eq:quasitransform}).
\end{defn}

\subsubsection{Example for single step quasi-symplectic Langevin flow}
\label{e.g.1}

We illustrate  below definition \ref{defflow} of  a quasi-symplectic Langevin normalizing flow in case of  a single transform  applied on a single random variable. We consider a probability measure $p(x)$ of random variable set $x \in X$. 
Then a single step Langevin flow transforms the original random variable $x$ to a new random variable $y = \Psi_0(x), y \in Y$. 
According to  definition \ref{defflow}, the new probability measure $q(y)$ of random variable $y$ is given by:
\begin{equation}
    q(y) = \mathcal{L}^0 p(x): \Psi_{0} \circ p(x) = p(\Psi_0^{-1}(y))
\end{equation}
By Eq.(\ref{eq:NFsF}), we conclude that:
\begin{equation}
    q(y) = p(x) \cdot |det \frac{\partial \Psi_0}{\partial x}|^{-1}
\end{equation}

The defined quasi-symplectic Langevin flow is a generalization of the Langevin  flow  with a quasi-symplectic structure for the parameters phase space. The quasi-symplectic Langevin  normalizing flow has  a deterministic kernel  $\Psi_0$  when the kernel stochastic factor $\sigma = 0$, and degenerates to a symplectic transition when $\nu = 0$. 

\section{Quasi-symplectic Langevin VAE}
\label{headings}

\subsection{Lower Bound Estimation With LVAE}
In the quasi-symplectic Langevin VAE, we use an augmented latent space consisting of position $\phi_I$ and velocity $\kappa_I$ variables of dimension $\zeta$ : $z=(\phi_I,\kappa_I)$. The objective of the autoencoder is to optimize its parameters as to maximize the evidence lower bound $\mathbb{\tilde{L}}$: 
\begin{equation}
\centering
\begin{split}
    &\log{p(x)} = \log{\int_\Omega p(x, z) dz}=\log{\int_\Omega \tilde{p}(x)q(\tilde{z}|x) d\tilde{z}}\\
    & \geq \int_\Omega \log{\tilde{p}(x)q(\tilde{z}|x)} d\tilde{z} \equiv \mathbb{\tilde{L}}
\end{split}
\end{equation}
where $\Omega$ is the measure space of the latent variables and as $\tilde{p}(x)$ is an unbiased estimator for $p(x)$. The lower bound is equal to the evidence when the posterior approximation is equal to the true posterior. 
Thus maximizing the lower bound is equivalent to   minimize the gap between the true posterior $p(z|x)$ and its approximation $q(z|x)$~\citep{VIsummaryBlei}. 

\begin{algorithm}[H]
\SetAlgoLined
\SetKwInput{KwData}{Inputs}
\SetKwInput{KwResult}{Output}
\KwData{Data $X$, Inference steps I, damping $\nu$, time step $t$, prior $q_{\omega_E}^{0}(\phi_{0})$}
\KwResult{Encoding and decoding parameters $\omega=(\omega_E,\omega_D)$}
 Initialize all parameters, variables\;
 Define: $K_{II}(t, p) = p e^{-\nu t}$\;
\While{NOT $\omega$ converged}{  
    Get minibatch: $X_N \xleftarrow[]{N} X$\; 
    \While{NOT j = N}{
    $x_j \xleftarrow[]{j}X_N$   \tcp*{Get $x_j$ in minibatch.}
    $\phi_0\sim q_{\omega_E}^{0}(\phi_{0}|x_j)\;$  \tcp*{Sampling latent variable from variational prior}
    $\kappa_0 \sim \mathcal{N}(0,E_\zeta) \;$  \tcp*{Sampling velocity from unit Gaussian.}
      \For{$i = 1;\ i < I;\ i++$}{
            \tcp{Quasi-symplectic Langevin Normalizing Flow}
            $\kappa_{1,i} \gets K_{II}(\frac{t}{2}, \kappa_i);$
            $\phi_{1,i} \gets \phi_i -t \frac{\partial \log( p(x,\phi_))}{2 \partial \phi_i}$;\\
            $\kappa_{i+1} \gets K_{II}(\frac{t}{2},\kappa_{1,i})$;
            $\phi_{i+1} \gets \phi_{1,i} + \frac{t}{2} \kappa_{1,i}$\;
            }
        $p^*_\omega  \gets  \hat{p}_{\omega_D}(x,\phi_{I}) \cdot \mathcal{N}(\kappa_I|0,E_\zeta)$\;
        $q^*_\omega  \gets  q_{\omega_E}^{0}(\phi_{0})\cdot \mathcal{N}(\kappa_0|0,E_\zeta) exp(\nu t)$\; 
        $\mathbb{\tilde{L}^{*}}_j \gets \log(p^*_\omega) - \log(q^*_\omega)$; \tcp*{Quasi-symplectic Langevin ELBO}
        $j \gets j + 1$
    }
    $\mathbb{\tilde{L}^{*}} \gets \sum_{i=1}^N \mathbb{\tilde{L}^{*}}_i / N$ \tcp*{Minibatch average ELBO}
    $\argmax_{\omega \in \mathbb{R}^n} \mathbb{\tilde{L}^{*}}$ \tcp*{Optimize average ELBO over parameters subset}
}
\label{alg1}
\caption{Quasi-symplectic Variational Inference}
\end{algorithm}
The posterior approximation $q(z)$ is computed through a series of Langevin transformations which is the Langevin flow: $q_{\omega_E}(z|x)=q^{I}(\mathcal{L}^{I}(\phi_{0},\kappa_0))=q_{\omega_E}^{0}(\phi_{0},\kappa_0|x)\prod_{i=1}^{I}|det \nabla{\Psi_0(\phi_{i},\kappa_i))}|^{-1}=q_{\omega_E}^{0}(\phi_{0},\kappa_0|x)\exp(I\nu t)$, where $q_{\omega_E}^{0}(\phi_{0},\kappa_0|x)$ is  an initial approximation  parameterized by $\omega_E$ which can also be seen as the prior on random variables $\phi_{0},\kappa_0$.  

Similarly to the HVAE~\citep{Caterini2018HamiltonianVA}, an estimator of $p(x)$ is given by: ${\tilde{p}(x)} = \frac{{\hat{p}(x, \mathcal{L}^{I}(\theta_{0}, k_0))}}{ {q_{\omega_E}^{0}(\theta_{0}, \kappa_0)}}$
We then give the lower bound for the quasi-symplectic Langevin VAE as :
\begin{equation}
\centering
\begin{split}
    &\mathbb{\tilde{L}}:= \int_\Omega q_{\omega_E}(\tilde{z}|x) \cdot (\log{\hat{p}(x, \mathcal{L}^{I}(\phi_{0}, k_0))} - log( q_{\omega_E}^{0}(\phi_{0}, k_0)) +  I\nu t d\tilde{z}
\end{split}
\label{eq:elboLangevin_Origin}
\end{equation}


\subsection{Quasi-symplectic Langevin VAE}
The quasi-symplectic Langevin lower bound $\mathbb{\tilde{L}}$ lays the ground  for the stochastic inference of a variational auto-encoder. Given a set of dataset $ X :\{x^i \in X ; i \in \mathbb{N}_{+}\}$, we aim to learn  a generative model of that dataset from a  latent space with the  quasi-symplectic Langevin inference. The generative model $p(x,z)$ consists of a prior on initial variables $z_0=(\phi_{0},\kappa_0)$,  $q_{\omega_E}^{0}(\phi_{0},\kappa_0|x)=q_{\omega_E}^{0}(\phi_0|x)\cdot\mathcal{N}(\kappa_0|0,I_{\zeta})$  and conditional likelihood $p_{\omega_D}(x|z)$ parameterized by $\omega_D$. 
The Gaussian unit prior $\mathcal{N}(\kappa_0|0,I_{\zeta})$   is the canonical velocity distribution from which the initial velocity of the Langevin diffusion will be performed. The distribution $q_{\omega_E}^{0}(\phi_0|x)$ is the variational prior that depends on the data $x^i$. Thus the generative model $p_{\omega_E,\omega_D}(x,z)$ is  parameterized by both encoders and decoders and the 
 quasi-symplectic Langevin lower bound writes as:
\begin{equation}
\centering
\begin{split}
     \argmax_{\omega_E,  \omega_D} \mathbb{\tilde{L}^{*}} = &\mathbb{E}_{\phi_0\sim q_{\omega_E}^{0}(), \kappa_0\sim \mathcal{N}_\zeta()} (\log{\hat{p}_{\omega_E,\omega_D}(x, \mathcal{L}^{I}(\phi_{0}, \kappa_0))} - \\ &\log(q_{\omega_E}^{0}(\phi_{0}, \kappa_0)) + K\nu t)
\end{split}
\label{eq:elboLangevinArgmax}
\end{equation}
The  maximization of  the lower bound (\ref{eq:elboLangevinArgmax}), is can be performed efficiently with the reparameterization trick depending on the choice of the variational prior $q_{\omega_E}^{0}(\phi_0)$. 
To have a fair comparison with prior work \citep{Caterini2018HamiltonianVA}, we also perform Rao-Blackwellization  for reducing the variance of the ELBO in the  quasi-symplectic Langevin VAE:
\begin{equation}
\centering
\begin{split}
    \argmax_{\omega_E,  \omega_D} \mathbb{\tilde{L}^{*}} = 
     &\mathbb{E}_{\phi_0\sim q_{\omega_E}^{0}(), \kappa_0\sim \mathcal{N}_\zeta()}(\log{\hat{p}_{\omega_E,\omega_D}(x, \mathcal{L}^{I}(\phi_{0},\kappa_0))} \\
     &- \log( \hat{q}_{\omega_E}(\phi_{0}, \kappa_0)) +K\nu t
     -\frac{1}{2}\kappa_I^T\kappa_I) + \frac{\zeta}{2}; ~~~~\forall \phi_0, \kappa_0 \in \mathbb{R}^\zeta
\end{split}
\label{eq:elboLangevinReducing}
\end{equation}
The resulting algorithm is described in Alg.\ref{alg1}.
\begin{figure}[]
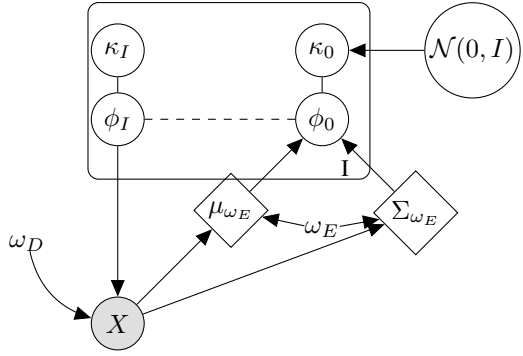

	\centering
\tikz{ %
    \node[obs, ] (X) {$X$} ; %
    \node[latent, above =2cm of X] (phi_i) {$\phi_I$} ; %
    \node[latent, right=2cm of phi_i] (phi) {$\phi_0$} ; %
    \node[det, below right=of phi] (Sig) {$\Sigma_{\omega_E}$} ; %
    \node[det, below left=of phi] (Mu) {$\mu_{\omega_E}$} ; %
    \node[const, below =of phi] (E) {$\omega_E$} ; %
    \node[latent, above=of phi, yshift=-8mm] (Kapa) {$\kappa_0$} ; %
        \node[latent, right =of Kapa] (Normal) {$\mathcal{N}(0,I)$} ; %
    \node[latent, above=of phi_i, yshift=-8mm] (Kapa_i) {$\kappa_I$} ; %
    \node[const, above left=of X] (od) {$\omega_D$} ; %
    \edge {Mu} {phi} ; %
    \edge {Sig} {phi} ; %
    \edge {Normal} {Kapa} ; %
     \edge {X} {Mu} ; %
     \edge {X} {Sig} ; %
      \edge {E} {Sig} ; %
       \edge {E} {Mu} ; %
    \edge[, bend right] {phi_i} {X} ; %
    \path (phi) edge[dashed,]  (phi_i) ;%
    \path (Kapa) edge[,]  (phi) ;%
    \path (Kapa_i) edge[,]  (phi_i) ;%
    \draw[->, to path={-| (\tikztotarget)}, bend right] (od) edge (X);
    \plate[inner sep=0.15cm, xshift=0.12cm, yshift=0.12cm] {plate1} {(phi) (phi_i) (Kapa) (Kapa_i)} {I}; %
  }
	\caption[]{Graphical model of the Quasi-symplectic Langevin Variational Autoencoder. The multivariate Gaussian parameters $\mu_{\omega_E}, \Sigma_{\omega_E}$  defining the variational prior of latent variable $\phi_0$ are  determined from the data $X$ and the parameter $\omega_E$ of the encoding network.  The initial velocity  latent variable $\kappa_0$ has a unit Gaussian prior and is paired by initial latent variable $\phi_0$. After iterating $I$ times the quasi-symplectic Langevin transform, the  latent pair $\{\phi_{I},\kappa_{I}\}$ is obtained from the initial variables $\{\phi_{0},\kappa_{0}\}$. The decoder  network with parameters $\omega_D$ is  then used to predict the data from latent variables $\phi_I$ through the conditional likelihood $p(x|\phi_I)$. Variables in diamonds are deterministically computed. Network parameters $\omega_E$, $\omega_D$ are optimized to maximize the ELBO.} 
	\label{fig:qslvaeflow}
\end{figure}
\section{Experiment and Result}
\label{others}
We examine the performance of quasi-symplectic Langevin VAE on the MNIST dataset \citep{lecun2010mnist} based on  various metrics.
\cite{Caterini2018HamiltonianVA} have reported that the Hamiltonian based stochastic variational inference outperforms that of Planar Normalizing Flow, mean-field based Variational Bayes in terms of model parameters inference error and quantitatively shown that the HVAE outperforms the naive VAE method in terms of  Negative Log-likelihood (NLL) score and ELBO. Here, we compare the proposed LVAE with the HVAE on MNIST dataset. The experiments were implemented with \textit{TensorFlow 2.0} and \textit{TensorFlow Probability} framework to evaluate the proposed approach in both qualitative and quantitative metrics. 

Given a training dataset  $X :\{x^i \in X ; i \in \mathbb{N}_{+}\}$ consisting of binary images of size $d$,  $x^i\in\{0,1\}^d$, we define the conditional likelihood $p(x|z)$ as a product of $d$ Bernoulli distributions. More precisely, we consider a decoder neural network $\decod_{\omega_D}(\phi)\in [0,1]^d$ that outputs $d$ Bernoulli parameters from the latent variable $\phi\in\mathbb{R}^\zeta$ where $z=(\phi,\kappa)$. Then the 
conditional likelihood writes as : $p(x^i|z^i)=\prod_{j=1}^d \decod_{\omega_D}(\phi)[j]^{x^i[j]}~  (1-\decod_{\omega_D}(\phi)[j])^{1-x^i[j]}$.

\begin{figure*}[!ht]
    \centering
    \includegraphics[width=\textwidth]{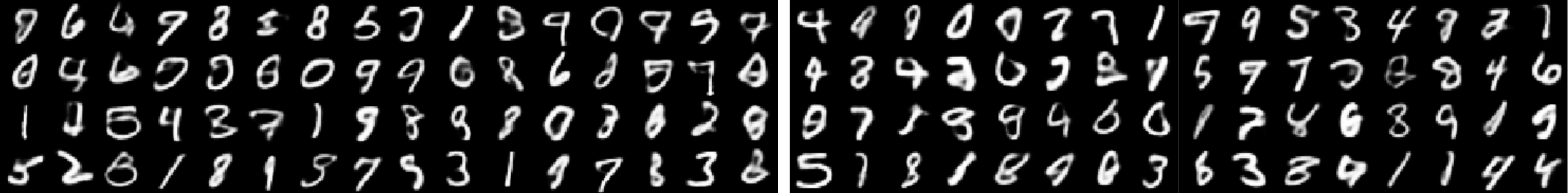}
    \caption{Quantitative result of Langevin VAE in comparison with HVAE. Left sub-figures are generated samples of HVAE. Right are samples of Langevin-VAE. In both  methods, the number of steps in the flow computation is $K=5$.}
    \label{fig:Quantitative_result}
\end{figure*}
\begin{table*}[ht]
\centering
\caption{Quantitative evaluation of the Langevin-VAE in comparison with the HVAE. It includes the comparison of the negative log likelihoods (NLL), the evidence lower bound (ELBO), the Fréchet Inception (FID) and Inception Score (IS)~\citep{FIDIS}}
\begin{tabular}{cccccccc}
           & \multicolumn{3}{c}{Langevin-VAE}        &  & \multicolumn{3}{c}{HVAE}           \\ 
\cline{2-4}\cline{6-8}
Flow steps & 1               & 5         &10      &  & 1               & 5        &10        \\ 
\hline
NLL        &\textbf{89.41} & \textit{\textbf{88.15}} &\textbf{89.63} &  &89.60& 88.21 & 89.69 \\
ELBO       &\textbf{-91.74}  & \textit{\textbf{-90.14} }&\textbf{-92.03} &  & -91.91  $\pm$ 0.01 & -90.41 &-92.39 $\pm$ 0.01  \\
FID        &\textit{\textbf{52.70}}     & \textbf{52.95}    &  \textbf{53.13}   &  &   53.12     &  53.26  & 53.21   \\
IS       & 6.42     & \textit{\textbf{6.49}} & \textbf{6.30}   &  & \textbf{6.57}   & 6.42 & 6.11\\
\hline
\end{tabular}
\label{tab:result}
\end{table*}
\subsection{Quasi-symplectic Langevin VAE on binary image benchmark}
\subsubsection{Implementation details}
Following the classical VAE approach~\citep{KingmaW13}, the encoder network parameterized by $\omega_E$ outputs  multivariate Gaussian parameters : $\mu_{\omega_E}(x)\in\mathbb{R}^\zeta$ and $\Sigma_{\omega_E}(x)\in\mathbb{R}^{\zeta}$, such that the variational prior is a multivariate Gaussian $q_{\omega_E}^{0}(\phi_{0}|x)=\mathcal{N}(\phi_{0}|\mu_{\omega_E}(x),\Sigma_{\omega_E}(x))$ with diagonal covariance matrix. This choice obviously makes the reparameterization trick feasible to estimate the lower bound. 
The related graphical model of the  quasi-symplectic Langevin VAE is displayed in Fig.~\ref{fig:qslvaeflow}.

The decoder and encoder neural network architectures are similar to the HVAE~\citep{Caterini2018HamiltonianVA} and MCMCVAE~\citep{salimans15}, both having three layers of 2D convolutional neural networks for encoder and decoder, respectively. The encoder network accepts a batch of data of size  $(N_b \times 28 \times 28)$ with $N_b = 1000$. The dimension of latent variables is set as $\zeta = 64$ and the damping factor is $\nu = 0$. The discretization step is $t = 1e-2$. The training stage stops when the computed ELBO does not improve on a validation dataset after 100 steps or when the inference loop achieves 2000 steps. 

Both  tested models LVAE and HVAE share the same training parameters.
The stochastic ascent of the ELBO is based on the  Adamax optimizer with a learning rate $lr = 5e-5$. All estimation of computation times were performed 
 on one NVIDIA  GeForce GTX 1080 Ti GPU. 

\subsubsection{Result on MNIST}
Both qualitative and quantitative results are studied. The generated samples of Langevin-VAE and HVAE are shown in Fig:~(\ref{fig:Quantitative_result}). We qualitatively see  that the  quality and  diversity of the sampled images 
are guaranteed for both autoencoder models. Quantitatively, Table~\ref{tab:result} shows the performance in terms of the NLL, ELBO, FID, IS scores  for Langevin-VAE and HVAE where Langevin and Hamiltonian flows  are experimentally compared.


\subsection{Quasi-symplectic Langevin VAE on Medical Image dataset}
We employ the proposed method for inference the cochlea dataset to generate the cochlea CT images. The human cochlea is an auditory nerve organ with a spiral shape. Some severe hearing impairments can be treated with cochlear transplantation. The shape of the cochlea is of great significance to the formulation of preoperative and postoperative plans. The quantitative analysis of the shape of the cochlea needs to process a large amount of CT image data. In this experiment, we use the proposed method to model the cochlear CT images dataset. 
\subsubsection{Dataset}
The dataset includes 1080 patients images that collected from the radiology department of *** University Hospital. The original slices sequences are with a spacing size of $0.185mm, 0.185mm, 0.25 mm$. We used a reference image to register all the images to the cochlea region (FOI) by using an automatic pyramidal blocking-matching (APBM) framework~\citep{Ourselin2000,Toussaint2007}. The FOI volumes are sampled into isometric spacing size of $0.2mm$ with 3D shape of $(60, 50, 50)$.
\subsubsection{Implementation details}
The proposed 3D Langevin-VAE consists with two 3D CNN. The encoder takes tensors with shape of $(N_b = 10 \cdot N_c = 60, N_w = 50, N_h = 50)$ and processes the tensors by three 3D convolutional layers with softplus non-linear projections. The strides of all the convolutional layers are set as 2. At the end, The tensors are flattened to fully connected layer for get the mean and variance parameters. The outputs parameters are then applied with the Langevin flow presented in Alogorithm \ref{alg1}~for adding the target information. The decoder network accepts the parameters and using the deconvolutional operation to model the marginal likelihood $p(x|z)$. The decoder follows inverse operations as the encoder did to upscale the feature maps to the original tensor shape. 
\subsubsection{Result on real dataset}
\begin{figure*}[!ht]
    \centering
    \includegraphics[width=\textwidth]{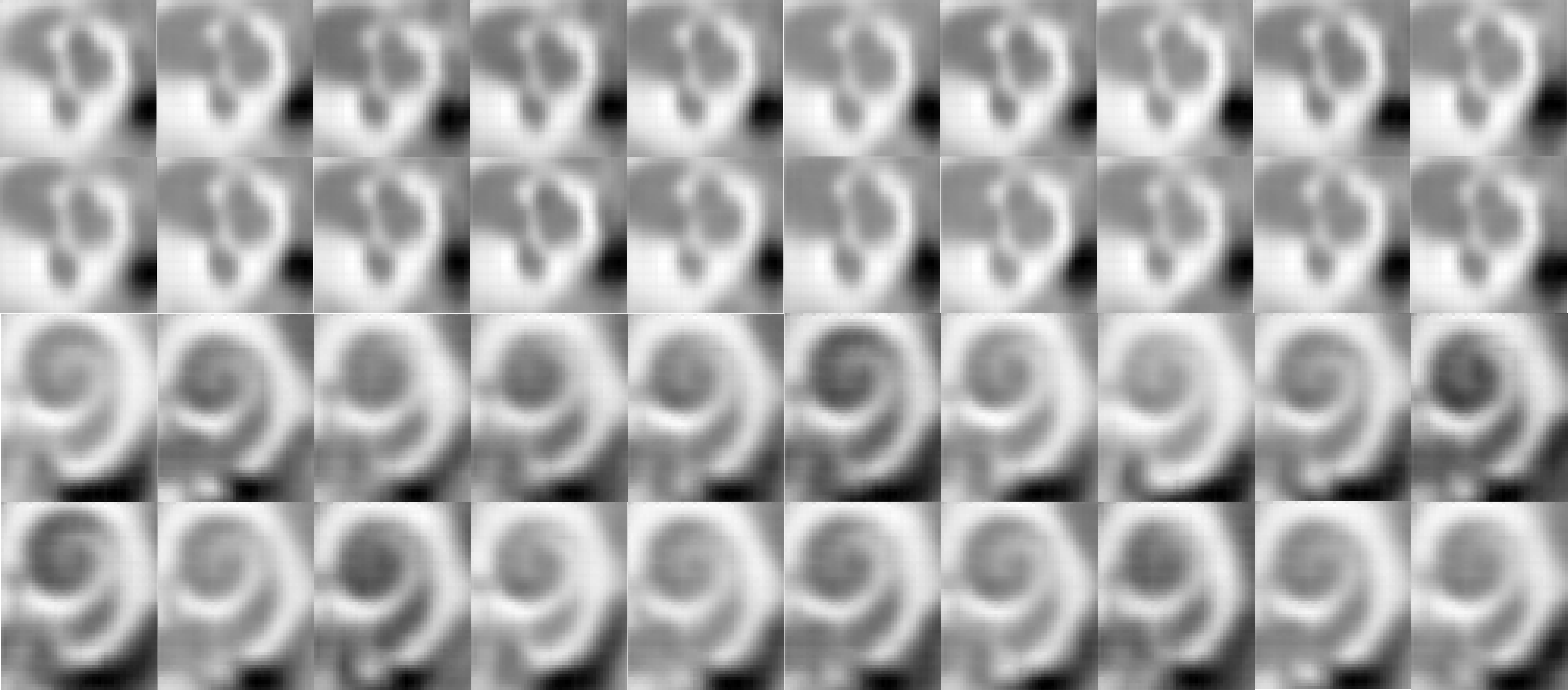}
    \caption{Qualitative assessment of the generated samples of Langevin-VAE. The number of steps in the flow computation is $K=5$.}
    \label{fig:result_cochlea}
\end{figure*}

Tab.~\ref{tab:comparison_vae} shows two inference metrics that represent the inference performance on the medical image dataset.  We see that the Langevin-VAE outperforms the VAE method on the dataset abstraction ability (as the ELBO and NLL are all better than VAE). We see that the Langevin-VAE can converge with 174 steps, while the VAE needs 201 steps for convergence. Fig.~\ref{fig:result_cochlea} shows 20 samples of generated fake cochlea CT images. We see that the Langevin-VAE learns the variance of the cochlea shapes and diversity of the intensity changes. 

\begin{table}
\centering
\caption{Quantitative evaluation of the Langevin-VAE in comparison with the VAE}
\label{tab:comparison_vae}
\begin{tabular}{ccc} 
\hline
                    & \textbf{VAE}                        & \textbf{Langevin-VAE}         \\
\textbf{Avg. ELBO}  & -85293.33 +/- 1.538 & \textbf{-85135.24 +/- 4.82 }  \\
\textbf{Avg. NLL}   & 83204.7+/- 10.92  & \textbf{83159.44 +/- 5.31 }   \\
\textbf{Early Stop} & 201                                 & \textbf{174 }                 \\
\hline
\end{tabular}
\end{table}


\section{Conclusion}
In this paper, we propose a new flow-based Bayesian inference framework by introducing the quasi-symplectic Langevin flow for the stochastic estimation of a tight ELBO. In comaprison with conventional VAE and HVAE, the proposed method achieves better performance on both toy and real world problems.
Specially, by introducing the quasi-symplectic Langevin dynamics, we also overcome the limitation of  the Langevin normalizing flow~\citep{gu} which requires to provide the Hessian matrix  $\nabla \nabla log(p(x,\phi))$ to compute the Jacobian. To the best of our knowledge, the proposed approach is the first Langevin flow based method as an generative model for dataset modeling.

Potential improvements of the quasi-symplectic Langevin inference can arise by investigating the manifold structure of the posterior  densities of the latent variables~\citep{Girolami,Alessandro,Livingstone2014InformationGeometricMC} to improve the inference efficiency.

\subsubsection*{Acknowledgments}
This work was partially funded by the French government through the UCA
JEDI ”Investments in the Future” project managed by the National Research
Agency (ANR) with the reference number ANR-15-IDEX-01, and was supported
by the grant AAP Santé 06 2017-260 DGA-DSH.

\bibliography{iclr2021_conference}
\bibliographystyle{iclr2021_conference}

\appendix
\section{Appendix}
\subsection{Over-damped form of the Generalized Langevin Diffusion}
\label{sec;overdamp}
We consider a unit mass $m=1$  evolving with a Brownian motion. 
The velocity part of the generalized Langevin type equation is:
\begin{eqnarray}
  \partial\Theta(t) = K dt
    \label{eq:velocity} & \partial K(t) = \frac{\partial\Theta(t)^2}{\partial t^2} = \frac{ \partial ln( p_{\Theta}(x))}{\partial \Theta}dt - \nu \Gamma K(t) + \delta \sigma(t)
\end{eqnarray}
In the case of an over-damped frictional force, the frictional force  $\nu K $  overwhelms the inertial force $m \cdot \partial^2 \theta/ \partial t^2$,   and thus $ \frac{\frac{\partial\Theta(t)^2}{\partial t^2} }{\nu  K(t)} \approx 0$.
According to the generalized Langevin diffusion equation, we have :
\[
\frac{\frac{\partial\Theta(t)^2}{\partial t^2} }{\nu  K(t)} = \frac{\frac{ \partial ln( p_{\Theta}(x))}{\partial \Theta}dt }{\nu  K(t)} -  \Gamma + \frac{\delta \sigma(t)}{\nu  K(t)}
\]
Therefore, we get :
\[
{\nu  K(t)}\Gamma  \approx \frac{ \partial ln( p_{\Theta}(x))}{\partial \Theta}dt + \delta \sigma(t) 
\]
which is the evolution given in Eq~\ref{qslangevin}. 

\subsection{Proof the integrator Eq.~\ref{eq:quasitransform} is quasi-symplectic}
\label{sec;symplectic}

\begin{prop}
\label{prop:quasi}
Eq.~\ref{eq:quasitransform} is asymptotic symplectic: $\lim_{\nu \to 0} |\Psi_0(K_i,\Phi_{i})| = exp(-\nu t)$
\end{prop}
\begin{remark}
Proposition [\ref{prop:quasi}] has propositional equivalences that the exterior power between two integration steps are equivalent as the Jacobian $|\Psi_0(K_i,\Phi_{i})|$ is not dependent on the time step term $t$. Thus, to prove the proposition \ref{prop:quasi} is equivalent to proof that: $dK_{i+1} \wedge d\Phi_{i+1} = dK_{i} \wedge d\Phi_{i}$.
\end{remark}
\textit{Proof:}

Let, $\nu \to 0$, the term $K_{II}$ of the composite integrator Eq.~\ref{eq:quasitransform} goes to: $\lim_{\nu \to 0} K_{II}(t,\phi) = \phi$

Then, 
\begin{equation}
\begin{split}
dK_{i+1} &= dK_{i} + t d(\frac{\partial \log(p(x,\Phi_{i} + \frac{t}{2} K_{i}))}{\partial \Phi_{i}})\\
&= dK_{i} + d[\frac{t \partial \log(p(x,\Phi_{i} + \frac{t}{2} K_{i}))}{\partial \Phi_{i}}](d\Phi_i +\frac{t}{2}K_i)\\
d\Phi_{i+1} &= d\Phi_{i} + \frac{t}{2}dK_{i} + \frac{t}{2}d(K_{i} + \frac{t \partial \log(p(x,\Phi_{i} + \frac{t}{2} K_{i})}{\partial \Phi_{i}})\\
&= d\Phi_{i} + tdK_{i} + d\frac{t^2\partial \log(p(x,\Phi_{i} + \frac{t}{2} K_{i})}{2\partial \Phi_{i}}(d\Phi_i +\frac{t}{2}K_i)
\end{split}
\label{eq:prove_sym}
\end{equation}
Let $U'= \frac{\partial \log(p(x,\Phi_{i} + \frac{t}{2} K_{i})}{2\partial \Phi_{i}}$, thus,

\begin{equation}
\begin{split}
dK_{i+1} \wedge d\Phi_{i+1} &= dK_{i} \wedge d\Phi_{i} + dK_{i} \wedge tdK_{i} + dK_{i} \wedge  \frac{t^2}{2}dU'd\Phi_{i} + \\
&dK_{i} \wedge \frac{t^3}{4}dU'dK_i + tdU'(d\Phi_i + \frac{t}{2} dK_i) \wedge d\Phi_i + tdU'(d\Phi_i + \\
&\frac{t}{2} dK_i) \wedge t dK_i + tdU'(d\Phi_i + \frac{t}{2} dK_i) \wedge \frac{t^2}{2}dU'(d\Phi_i + \frac{t}{2} dK_i)
\end{split}
\label{eq:simplified}
\end{equation}
According the property of exterior product, therefore:
\begin{equation}
\begin{split}
dK_{i} \wedge tdK_{i} = tdU'd\Phi_i \wedge d\Phi_i = tdU'\frac{t}{2} dK_i \wedge t dK_i = 0
\end{split}
\end{equation}
Simplifying Eq.~\ref{eq:simplified}:
\begin{equation}
\begin{split}
dK_{i+1} \wedge d\Phi_{i+1} &= dK_i \wedge d\Phi_i + t^2 dU'(dK_i \wedge d\Phi_i) + t^2 dU'(d\Phi_i \wedge dK_i) + \\
&\frac{t^4dU'^2}{4}(d\Phi_i \wedge dK_i) + \frac{t^4dU'^2}{4}(dK_i \wedge d\Phi_i)\\
&= dK_i \wedge d\Phi_i + t^2 dU'(dK_i \wedge d\Phi_i) - t^2 dU'(dK_i \wedge d\Phi_i) + \\
&\frac{t^4dU'^2}{4}(d\Phi_i \wedge dK_i) - \frac{t^4dU'^2}{4}(d\Phi_i \wedge dK_i)\\
&= dK_i \wedge d\Phi_i 
\end{split}
\label{eq:simplified2}
\end{equation}
Q.E.D.

\subsection{Evidence lower bound of Langevin Flow}
\label{sec;elbo}
We consider the log-likelihood: $\log{p(x)}$ with latent variables $z$, based on Jensen's inequality:
\begin{equation}
\centering
\begin{split}
    &\log{p(x)} \geq \int_\Omega \log{\tilde{p}(x)q(\tilde{\tilde{z}}|x)} d\tilde{z} 
\end{split}
\label{eq:elbo}
\end{equation}
The data prior is given through the Langevin flow where $\mathcal{L}^{I}(\theta_{0}, k_0)$ are the $K$ steps Langevin flows with initialization  states $(\theta_{0}, k_0)$:
\begin{equation}
\centering
\begin{split}
    &{\tilde{p}} = \frac{{\hat{p}(x, \mathcal{L}^{I}(\theta_{0}, k_0))}}{ {q^{0}(\mathcal{L}^{0}(\theta_{0}, k_0))}}
\end{split}
\end{equation}
Therefore, we can get the Langevin flow lower bound:
\begin{equation}
\centering
\begin{split}
     &\mathbb{\tilde{L}}: \geq \int_\Omega q(\tilde{z}|x) \cdot (\log{\hat{p}(x, \mathcal{L}^{I}(\theta_{0}, k_0))}- \log{q^{0}(\mathcal{L}^{0}(\theta_{0}, k_0))} )d\tilde{z} \\
     &=\int_\Omega q(\tilde{z}|x) \cdot (\log{\hat{p}(x, \mathcal{L}^{I}(\theta_{0}, k_0))} - \log( q^{0}(\theta_{0}, k_0)\prod_{k=1}^{I}|det \nabla{\Psi^{-1}_i{I}(\theta_{0}, k_0))|}))d\tilde{z}\\
     &=\int_\Omega q(\tilde{z}|x) \cdot (\log{\hat{p}(x, \mathcal{L}^{I}(\theta_{0}, k_0))} - \log( q^{0}(\theta_{0}, k_0)) - \sum_{k=1}^{I}\log(|det \nabla{\Psi^{-1}_{I}(\theta_{0}, k_0))|})) d\tilde{z}\\
     &=\int_\Omega q(\tilde{z}|x) \cdot (\log{\hat{p}(x, \mathcal{L}^{I}(\theta_{0}, k_0))} - \log( q^{0}(\theta_{0}, k_0)) + \sum_{k=1}^{I} (\nu t)) d\tilde{z}
\end{split}
\label{eq:elboLangevin}
\end{equation}


\end{document}